\documentclass[twoside]{article}

\usepackage[accepted]{aistats2017submitted}
\usepackage{natbib}
\usepackage{amsmath,amsfonts,amssymb,amsthm}
\usepackage{mathtools}
\mathtoolsset{showonlyrefs}
\usepackage{graphicx} 
\usepackage{subcaption}
\usepackage{macros}
\usepackage[font=small]{caption}
\usepackage[inline]{enumitem}
\usepackage{todonotes}
\begin{document}

%

%
\runningauthor{Linderman, Miller, Adams, Blei, Paninski, and Johnson}

\twocolumn[

\aistatstitle{Recurrent Switching Linear Dynamical Systems}

\aistatsauthor{ Scott W. Linderman \And Andrew C. Miller \And Ryan P. Adams }

\aistatsaddress{ Columbia University \And Harvard University \And Harvard University and Twitter } 

\vspace{-1em}
\aistatsauthor{ David M. Blei \And Liam Paninski \And Matthew J. Johnson }

\aistatsaddress{ Columbia University \And Columbia University \And Harvard University }
]

\begin{abstract}
Many natural systems, such as neurons firing in the brain or
basketball teams traversing a court, give rise to time series data
with complex, nonlinear dynamics.  We can gain insight into these
systems by decomposing the data into segments that are each explained
by simpler dynamic units.  Building on switching linear dynamical
systems (SLDS), we present a new model class that not only discovers
these dynamical units, but also explains how their switching behavior
depends on observations or continuous latent states.  These ``recurrent'' switching
linear dynamical systems provide further insight by discovering the
conditions under which each unit is deployed, something that
traditional SLDS models fail to do.  We leverage recent algorithmic
advances in approximate inference to make Bayesian inference in these
models easy, fast, and scalable.
\end{abstract}

\section{Introduction}
Complex dynamical behaviors can often be broken down into simpler units.
A basketball player finds the right court position and starts a pick and roll play.
A mouse senses a predator and decides to dart away and hide.
A neuron's voltage first fluctuates around a baseline until a threshold is exceeded; it spikes to peak depolarization, and then returns to
baseline.
In each of these cases, the switch to a new mode of behavior can depend on the
continuous state of the system or on external factors.
By discovering these behavioral units and their switching dependencies, we can
gain insight into complex data-generating processes.

This paper proposes a class of recurrent state space models that captures these
dependencies and a Bayesian inference and learning algorithm that is
computationally tractable and scalable to large datasets.
We extend switching linear-Gaussian dynamical systems (SLDS)
\citep{ackerson1970state, chang1978state,
  hamilton1990analysis, 
  ghahramani1996switching,
  murphy1998switching, fox2009nonparametric} by introducing a class of
models in which the discrete switches can depend on the continuous
latent state and exogenous inputs through a logistic regression.
Previous models including this
dependence, like the piecewise affine framework for hybrid dynamical systems \citep{sontag1981nonlinear},
abandon the conditional linear-Gaussian
structure in the continuous states and thus greatly complicate inference \citep{paoletti2007identification, juloski2005bayesian}.
Our main technical contribution is a new inference algorithm that leverages
auxiliary variable methods \citep*{polson2013bayesian,
  linderman2015dependent} to make inference both fast and easy.

The class of models and the corresponding learning and inference
algorithms  we develop have several advantages for understanding rich time series data.
First, these models decompose data into simple segments and
attribute segment transitions to changes in latent state or environment;
this provides interpretable representations of data dynamics. 
Second, we fit these models using fast, modular Bayesian inference
algorithms; this makes it easy to handle missing data, multiple observation
modalities, and hierarchical extensions.
Finally, these models are interpretable, readily able to incorporate prior
information, and generative; this lets us take advantage of a variety of
tools for model validation and checking.

In the following section we provide background on the key models and
inference techniques on which our method builds.
Next, we introduce the class of recurrent switching state space models, and
then explain the main algorithmic contribution that enables fast learning and
inference.
Finally, we illustrate the method on synthetic data experiments
and an application to recordings of professional basketball players.

\section{Background}
Our model has two main components: switching linear dynamical systems and
stick-breaking logistic regression.
Here we review these components and fix the notation we will use throughout the paper.

\subsection{Switching linear dynamical systems}
\label{sec:slds}
Switching linear dynamical system models (SLDS) break down complex, nonlinear
time series data into sequences of simpler, reused dynamical modes.
By fitting an SLDS to data, we not only learn a flexible nonlinear generative
model, but also learn to parse data sequences into coherent discrete units.

The generative model is as follows. At each time ${t=1,2,\ldots,T}$ there is a discrete latent state ${z_t \in \{1,2,
\ldots,K\}}$ that following Markovian dynamics,
\begin{equation}
  z_{t+1} \given z_t, \{\pi_k\}_{k=1}^K  \sim \pi_{z_t}
  \label{eq:markov_z}
\end{equation}
where ${\{\pi_k\}_{k=1}^K}$ is the Markov transition matrix and
${\pi_k \in [0,1]^K}$ is its $k$th row. 
In addition, a continuous latent state ${x_t \in \reals^M}$ follows
conditionally linear (or affine) dynamics, where the discrete state $z_t$
determines the linear dynamical system used at time $t$:
\begin{align}
  x_{t+1} &= A_{z_{t+1}} x_{t} + b_{z_{t+1}} +  v_t,
  &
  v_t &\iid\sim \mcN(0,Q_{z_{t+1}}),
  \quad
  \label{eq:slds_start}
\end{align}
for matrices ${A_k, Q_k \in \reals^{M \times M}}$ and vectors~${b_k\in
\reals^M}$ for~${k=1,2,\ldots,K}$.
Finally, at each time $t$ a linear Gaussian observation $y_t \in \reals^N$ is
generated from the corresponding latent continuous state,
\begin{align}
  y_t &= C_{z_t} x_t + d_{z_t} + w_t, & w_t &\iid\sim \mcN(0,S_{z_t}),
    \label{eq:slds_end}
\end{align}
for $C_k \in \reals^{N \times M}$, $S_k \in \reals^{N \times N}$,
and~$d_k \in \reals^N$.
The system parameters comprise the discrete Markov transition matrix and the
library of linear dynamical system matrices, which we write as
\begin{equation*}
  \theta = \{(\pi_k, A_k, Q_k, b_k, C_k, S_k, d_k)\}_{k=1}^K.
\end{equation*}
For simplicity, we will require~$C$,~$S$,
and~$d$ to be shared among all discrete states in our experiments.
In general, equations~\eqref{eq:slds_start} and \eqref{eq:slds_end} can be extended to include 
linear dependence on exogenous inputs,~${u_t \in \reals^P}$, as well.

To learn an SLDS using Bayesian inference, we place conjugate Dirichlet priors
on each row of the transition matrix and conjugate matrix normal
inverse Wishart (MNIW) priors on the linear dynamical system parameters,
writing
\begin{gather*}
  \pi_k \given \alpha \iid\sim \distDirichlet(\alpha),
  \quad
  (A_k, b_k), Q_k \given \lambda \iid\sim \distMNIW(\lambda),
  \\
  (C_k, d_k), S_k \given \eta \iid\sim \distMNIW(\eta),
\end{gather*}
where $\alpha$, $\lambda$, and~$\eta$ denote hyperparameters.

\begin{figure*}[t]
  \centering
  \includegraphics[width=6.5in]{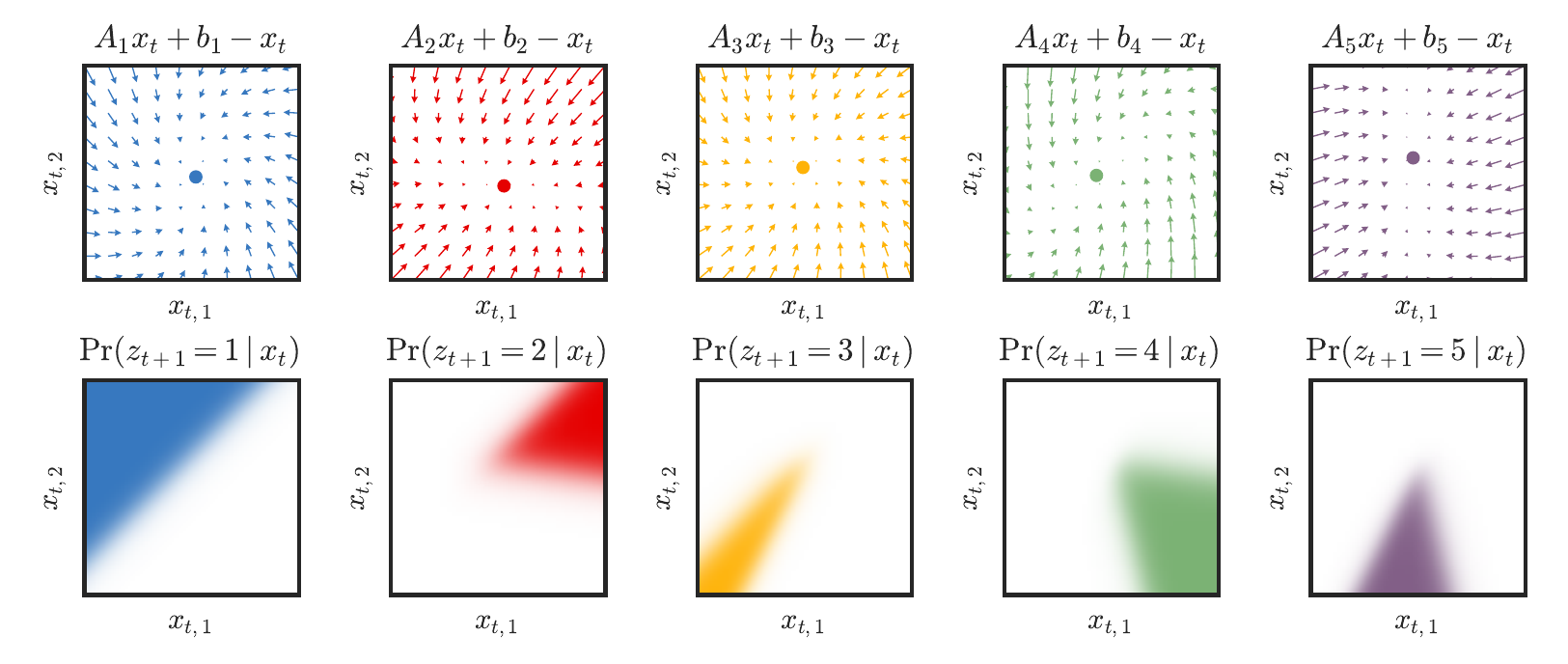}
  \vspace{-1em}
  \caption{A draw from the prior over recurrent switching
    linear dynamical systems with ${K=5}$ discrete states
    shown in different colors.
    {\bf(Top)} The linear dynamics of each latent state. Dots show
    the fixed point~${(I-A_k)^{-1} b_k}$.
    {\bf(Bottom)} The conditional $p(z_{t+1} | x_t)$ plotted as a function of
    $x_t$ (white=0; color=1). Note that the stick breaking construction iteratively partitions the
    continuous space with linear hyperplanes.
    For simpler plotting, in this example we restrict $p(z_{t+1} \given x_t, z_t) =
    p(z_{t+1} | x_t)$.
    }
  \label{fig:prior}
\end{figure*}

\subsection{Stick-breaking logistic regression and P\'olya-gamma augmentation}
\label{sec:stickbreaking}
Another component of the recurrent SLDS is a stick-breaking logistic regression,
and for efficient block inference updates we leverage a recent P\'olya-gamma
augmentation strategy \citep*{linderman2015dependent}.
This augmentation allows certain logistic regression evidence potentials to
appear as conditionally Gaussian potentials in an augmented distribution, which
enables our fast inference algorithms.

Consider a logistic regression model from regressors~${x \in \reals^M}$ to a
categorical distribution on the discrete variable~${z \in \{1,2,\ldots,K\}}$,
written as
\begin{equation}
  z \given x \sim \pisb(\nu),
  \quad
  \nu = R x + r,
\end{equation}
where ${R \in \reals^{K-1 \times M}}$ is a weight matrix and~${r \in \reals^{K-1}}$
is a bias vector.
Unlike the standard multiclass logistic regression, which uses a softmax link
function, we instead use a stick-breaking link function
${\pisb: \reals^{K-1} \to [0,1]^K}$, which maps a real vector to a
normalized probability vector via the stick-breaking process
\begin{gather*}
  \pisb(\nu) = \begin{pmatrix} \pisb^{(1)}(\nu) & \cdots & \pisb^{(K)}(\nu) \end{pmatrix},
  \\
  \pisb^{(k)}(\nu) = \sigma(\nu_k) \prod_{j < k} (1-\sigma(\nu_j))
  = \sigma(\nu_k) \prod_{j < k} \sigma(-\nu_j),
\end{gather*}
for ${k=1,2,\ldots,K-1}$ and $\pisb^{(K)}(\nu) =
\prod_{k=1}^K \sigma(-\nu_k)$,
where ${\sigma(x) = e^x / (1 + e^x)}$ denotes the logistic function.
The probability mass function~$p(z \given x)$ is
\begin{equation*}
  p(z \given x) = \prod_{k=1}^K \sigma(\nu_k)^{\bbI[z = k]} \sigma(-\nu_k)^{\bbI[z > k]}
\end{equation*}
where $\bbI[\,\cdot\,]$ denotes an indicator function that takes value $1$ when its
argument is true and $0$ otherwise.

If we use this regression model as a likelihood~$p(z \given x)$ with a
Gaussian prior density~$p(x)$, the posterior density~$p(x \given z)$ is
non-Gaussian and does not admit easy Bayesian updating.
However, \citet*{linderman2015dependent} show how to introduce P\'olya-gamma
auxiliary variables~${\omega = \{\omega_k\}_{k=1}^K}$ so that the conditional
density~$p(x \given z, \omega)$ becomes Gaussian.
In particular, by choosing~
${\omega_k \given x, z \sim \distPolyaGamma(\bbI[z \geq
k], \nu_k)}$, we have,
\begin{equation*}
  x \given z, \omega \sim \mcN(\Omega^{-1} \kappa, \; \Omega^{-1}),
\end{equation*}
where the mean vector~$\Omega^{-1} \kappa$ and covariance matrix~$\Omega^{-1}$
are determined by
\begin{align}
  \Omega &= \diag(\omega),
  &
  \kappa_k &= \bbI[z = k] - \frac{1}{2} \bbI[z \geq k].
\end{align}
Thus instantiating these auxiliary variables in a Gibbs sampler or
variational mean field inference algorithm enables efficient block updates
while preserving the same marginal posterior distribution $p(x \given z)$.


\section{Recurrent Switching State Space Models}
\label{sec:models}
The discrete states in the ``classical'' SLDS of Section~\ref{sec:slds} are generated via an \emph{open loop}: the discrete state $z_{t+1}$ is a function only of the
preceding discrete state~$z_t$, and~$z_{t+1} \given z_t$ is independent of the
continuous state~$x_t$.
That is, if a discrete switch should occur whenever the continuous state enters a
particular region of state space, the SLDS will be unable to learn this dependence.

\begin{figure*}[t]
  \centering
  \begin{subfigure}[t]{0.47\textwidth}
    \centering
    \includegraphics[width=\textwidth]{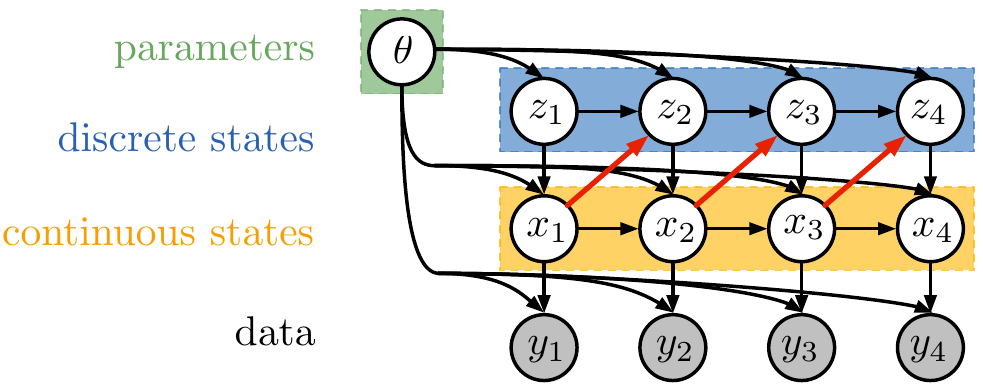}
    \caption{rSLDS}
    \label{fig:rslds}
  \end{subfigure}
  \hfill
  \begin{subfigure}[t]{0.47\textwidth}
    \centering
    \includegraphics[width=\textwidth]{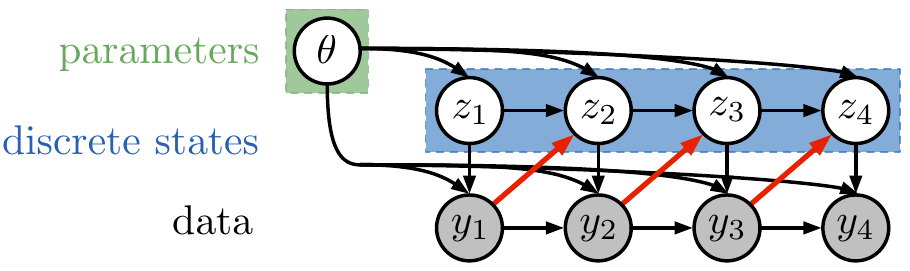}
    \caption{rAR-HMM}
    \label{fig:rarhmm}
  \end{subfigure}
  \caption{
  Graphical models for the recurrent SLDS (rSLDS) and recurrent AR-HMM
  (rAR-HMM).
  Edges that represent recurrent dependencies of discrete
  states on continuous observations or continuous latent states are
  highlighted in red.
  }
  \label{fig:pgms}
  \vspace{-1em}
\end{figure*}

We introduce the \emph{recurrent switching linear dynamical system} (rSLDS), an
extension of the SLDS to model these dependencies directly.
Rather than restricting the discrete states to open-loop Markovian dynamics as
in Eq.~\eqref{eq:markov_z}, the rSLDS allows the discrete state transition
probabilities to depend on additional covariates, and in particular on the
preceding continuous latent state.
That is, the discrete states of the rSLDS are generated as
\begin{align}
  z_{t+1} \given z_t, x_t, \{R_k, r_k \}  &\sim \pisb(\nu_{t+1}), \nonumber \\
  \nu_{t+1} &= R_{z_t} x_t + r_{z_t},
  \label{eq:nu}
\end{align}
where ${R_k \in \reals^{K-1 \times M}}$ is a weight matrix that specifies
the recurrent dependencies and~${r_k \in \reals^{K-1}}$ is a bias
that captures the Markov dependence of $z_{t+1}$ on $z_t$.
The remainder of the rSLDS generative process follows that of the SLDS from
Eqs.~\eqref{eq:slds_start}-\eqref{eq:slds_end}.
See Figure~\ref{fig:rslds} for a graphical model, where the edges representing
the new dependencies of the discrete states on the continuous latent states are
highlighted in red. As with the standard SLDS, both the discrete and continuous rSLDS
dynamics could be extended with linear dependence on exogenous inputs,~$u_t$, as well.

Figure~\ref{fig:prior} illustrates an rSLDS with~${K=5}$
discrete states and~${M=2}$ dimensional continuous states.
Each
discrete state corresponds to a set of linear dynamics defined
by~$A_k$ and~$b_k$, shown in the top row. The transition
probability,~$\pi_t$, is a function of the previous states~$z_{t-1}$
and~$x_{t-1}$. We show only the dependence on~$x_{t-1}$ in the bottom
row. Each panel shows the conditional probability,~${\Pr(z_{t+1}=k
  \given x_{t})}$, as a colormap ranging from zero (white) to
one (color). Due to the logistic stick breaking, the latent space is
iteratively partitioned with linear hyperplanes.

There are several useful special cases of the rSLDS.

\paragraph{Recurrent ARHMM (rAR-HMM)} Just as the autoregressive HMM (AR-HMM)
is a special case of the SLDS in which we observe the states~$x_{1:T}$
directly, we can define an analogous~rAR-HMM.
See Figure~\ref{fig:rarhmm} for a graphical model, where the edges
representing the dependence of the discrete states on the continuous observations
are highlighted in red.

\paragraph{Shared rSLDS (rSLDS(s))} Rather than having separate recurrence
weights (and hence a separate partition) for each value of~$z_t$, we can
share the recurrence weights as,
\begin{equation*}
  \nu_{t+1} = R x_{t} + r_{z_t}.
\end{equation*}

\paragraph{Recurrence-Only (rSLDS(ro))}There is no dependence on~$z_{t}$ in
this model. Instead,
\begin{equation*}
  \nu_{t+1} = R x_t + r.
\end{equation*}
While less flexible, this model is eminently interpretable, 
easy to visualize, and, as we show in experiments, works well for
many dynamical systems.
The example in Figure~\ref{fig:prior} corresponds to this special case.

\paragraph{Standard SLDS}
We can recover the standard SLDS with $\nu_{t+1} = r_{z_t}$.

\paragraph{Recurrent Sticky SLDS}
Rather than allowing the continuous states to govern the entire
distribution over next discrete states, $x_t$ may only affect
whether or not to stay in the current state. That is,
\begin{align}
  s_{t+1} &\sim \distBernoulli(\sigma(r_{z_t}^\trans x_t)),
  &
  z_{t+1} &\sim
  \begin{cases}
    \delta_{z_t} & \text{if } s_{t+1} = 1, \\
    \widetilde{\pi}_{z_t} & \text{o.w.}
  \end{cases}
\end{align}
where~${\widetilde{\pi}_k \in [0,1]^{K-1}}$ is a distribution on
states \emph{other than} the current state. The ``stay-or-leave''
variable,~$s_{t+1}$, depends on the current location.

\section{Bayesian Inference}
\label{sec:inference}
Adding the recurrent dependencies from the latent continuous states to
the latent discrete states introduces new inference challenges.
While block Gibbs sampling in the standard SLDS can be accomplished with
message passing because $x_{1:T}$ is conditionally Gaussian given~$z_{1:T}$
and~$y_{1:T}$, the dependence of~$z_{t+1}$ on~$x_t$ renders the recurrent SLDS
non-conjugate.
To develop a message-passing procedure for the rSLDS, we first review standard
SLDS message passing, then show how to leverage a P\'olya-gamma augmentation
along with message passing to perform efficient Gibbs sampling in the rSLDS.
We discuss stochastic variational inference \citep{hoffman2013stochastic}
in the supplementary material.

\subsection{Message Passing}
First, consider the conditional density of the latent continuous state
sequence $x_{1:T}$ given all other variables, which is proportional to
\begin{equation}
  \prod_{t=1}^{T-1} \psi(x_t, x_{t+1}, z_{t+1}) \,
  \psi(x_t, z_{t+1})
  \prod_{t=1}^T \psi(x_t, y_t),
\end{equation}
where $\psi(x_t, x_{t+1}, z_{t+1})$ denotes the potential from the
conditionally linear-Gaussian dynamics and $\psi(x_t, y_t)$ denotes the
evidence potentials.
The potentials $\psi(x_t, z_{t+1})$ arise from the new dependencies in the
rSLDS and do not appear in the standard SLDS.
This factorization corresponds to a chain-structured undirected graphical model
with nodes for each time index.

We can sample from this conditional distribution using message passing.
The message from time $t$ to time $t'=t+1$, denoted $m_{t \to t'}(x_{t'})$, is
computed via
\begin{equation}
  \int \psi(x_t, y_t) \psi(x_t, z_{t'}) \psi(x_t, x_{t'}, z_{t'}) 
  m_{t'' \to t}(x_t) \diff x_t,
  \label{eq:message}
\end{equation}
where $t''$ denotes $t-1$.
If the potentials were all Gaussian, as is the case without the
rSLDS potential $\psi(x_t, z_{t+1})$, this integral could be computed
analytically.
We pass messages forward once, as in a Kalman filter, and then sample
backward.
This constructs a joint sample $\hat{x}_{1:T} \sim p(x_{1:T})$ in~$O(T)$ time.
A similar procedure can be used to jointly sample the discrete state
sequence,~$z_{1:T}$, given the continuous states and parameters.
However, this computational strategy for sampling the latent continuous states
breaks down when including the non-Gaussian rSLDS potential $\psi(x_t,
z_{t+1})$.

Note that it is straightforward to handle missing data in this formulation;
if the observation~$y_t$ is omitted, we simply have one fewer
potential in our graph.

\subsection{Augmentation for non-Gaussian Factors}
\label{sec:augmentation}
The challenge presented by the recurrent SLDS is that~$\psi(x_t,
z_{t+1})$ is not a linear Gaussian factor; rather, it is a categorical
distribution whose parameter depends nonlinearly on~$x_t$. Thus, the
integral in the message computation \eqref{eq:message} is not
available in closed form.  There are a number of methods of
approximating such integrals, like particle filtering
\citep{doucet2000sequential} or Laplace approximations \citep{tierney1986accurate}, but
here we take an alternative approach using the recently developed
\polyagamma augmentation scheme~\citep{polson2013bayesian}, which
renders the model conjugate by introducing an auxiliary variable in
such a way that the resulting marginal leaves the original model
intact.

According to the stick breaking transformation described in
Section~\ref{sec:stickbreaking}, the non-Gaussian factor is
\begin{equation*}
  \psi(x_t, z_{t+1}) =
  \prod_{k=1}^K \sigma(\nu_{t+1, k})^{\bbI[z_{t+1}=k]} \,
  \sigma(-\nu_{t+1,k})^{\bbI[z_{t+1}>k]},
\end{equation*}
where~$\nu_{t+1,k}$ is the~$k$-th dimension of~$\nu_{t+1}$, as defined
in~\eqref{eq:nu}. Recall that~$\nu_{t+1}$ is linear in~$x_t$.  
Expanding the definition of the logistic function, we have,
\begin{equation}
  \psi(x_t, z_{t+1}) =
  \prod_{k=1}^{K-1} \frac{(e^{\nu_{t+1,k}})^{\bbI[z_{t+1}=k]}}
       {(1+e^{\nu_{t+1,k}})^{\bbI[z_{t+1}\geq k]}}.
  \label{eq:psixz}
\end{equation}

The P\'{o}lya-gamma augmentation targets exactly such densities,
leveraging the following integral identity:
\begin{equation}
\label{eq:pg_identity}
\frac{(e^{\nu})^a}{(1+e^{\nu})^b} = 2^{-b} e^{\kappa \nu} \int_{0}^{\infty} e^{-\omega \nu^2 /2} p_{\distPolyaGamma}(\omega \given b, 0) \, \mathrm{d}\omega,
\end{equation}
where~${\kappa=a-b/2}$ and~$p_{\distPolyaGamma}(\omega\given b, 0)$ is the density of the P\'{o}lya-gamma
distribution,~${\distPolyaGamma(b, 0)}$, which does not depend on $\nu$.

Combining~\eqref{eq:psixz} and~\eqref{eq:pg_identity}, we see
that~$\psi(x_t, z_{t+1})$ can be written as a marginal of a factor on
the augmented space,~$\psi(x_t, z_{t+1}, \omega_t)$, where~${\omega_t
  \in \reals_+^{K-1}}$ is a vector of auxiliary variables.
As a function of~$\nu_{t+1}$, we have
\begin{equation}
  \psi(x_t, z_{t+1}, \omega_t) \propto 
  \prod_{k=1}^{K-1}
  \exp \big\{ \kappa_{t+1,k} \, \nu_{t+1,k}
  - \tfrac{1}{2}\omega_{t,k} \, \nu_{t+1,k}^2 \big\},
\end{equation}
where~${\kappa_{t+1,k} = \bbI[z_{t+1}=k] - \tfrac{1}{2}\bbI[z_{t+1} \geq k]}$.
Hence,
\begin{equation}
  \psi(x_t, z_{t+1}, \omega_t) \propto \distNormal(\nu_{t+1} \given \Omega_t^{-1} \kappa_{t+1}, \, \Omega_t^{-1}),
\end{equation}
with~${\Omega_t = \diag(\omega_t)}$
and~${\kappa_{t+1}=[\kappa_{t+1,1}\ldots, \kappa_{t+1,K-1}]}$.  Again,
recall that~$\nu_{t+1}$ is a linear function of~$x_t$.  Thus, after
augmentation, the potential on~$x_t$ is effectively Gaussian and the
integrals required for message passing can be written
analytically. Finally, the auxiliary variables are easily updated as
well, since~${\omega_{t,k} \given x_t, z_{t+1} \sim
  \distPolyaGamma(\bbI[z_{t+1} \geq k], \nu_{t+1, k})}$.

\subsection{Updating Model Parameters}
Given the latent states and observations, the model parameters benefit
from simple conjugate updates. The dynamics parameters have conjugate
MNIW priors, as do the emission parameters. The recurrence weights are
also conjugate under a MNIW prior, given the auxiliary
variables~$\omega_{1:T}$. We set the hyperparameters of these priors
such that random draws of the dynamics are typically stable and have
nearly unit spectral radius in expectation, and we set the mean of the
recurrence bias such that states are equiprobable in expectation.
We will discuss initialization in the following section.

\subsection{Initialization}
Given the complexity of
these models, it is important to initialize the parameters and latent
states with reasonable values. We used the following initialization
procedure:
\begin{enumerate*}[label=(\roman*)]
\item use probabilistic PCA or factor analysis to initialize the continuous
  latent states,~$x_{1:T}$, and the observation,~$C$,~$S$, and~$d$;
\item fit an AR-HMM to~$x_{1:T}$ in order to initialize the discrete latent
  states,~$z_{1:T}$, and the dynamics models,~$\{A_k, Q_k, b_k\}$; and
  then
\item greedily fit a decision list with logistic regressions at each node
  in order to determine a permutation of the latent states most amenable
  to stick breaking. 
\end{enumerate*}
In practice, the last step alleviates the undesirable dependence on ordering
that arises from the stick breaking formulation. We discuss this and alternative approaches
in more depth in the supplementary material.

With this initialization, the Gibbs sampler refines the parameter and
state estimates and explores (at least a mode of) the posterior.

\begin{figure*}[t]
  \centering
  \includegraphics[width=6.5in]{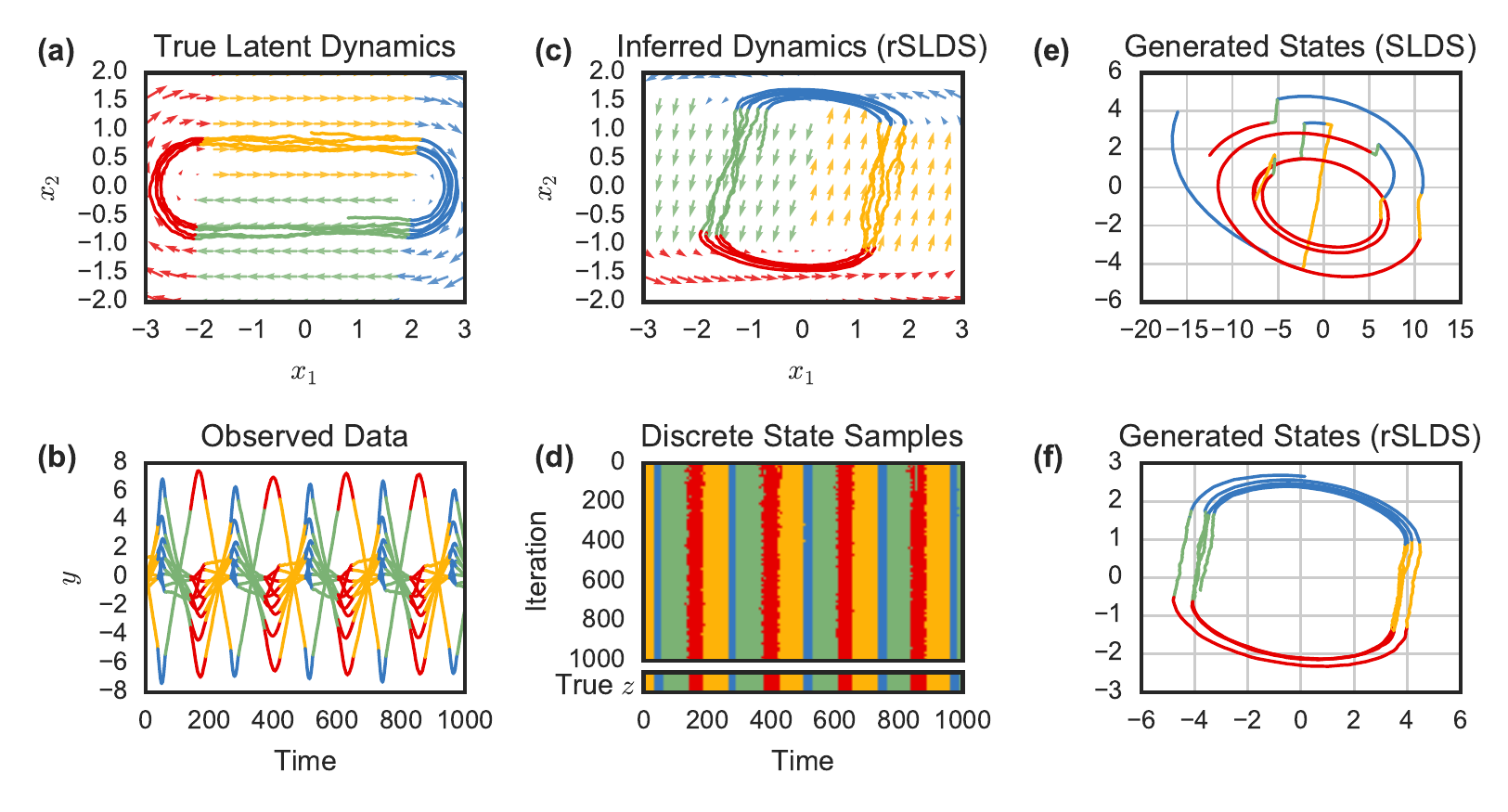}
  \vspace{-1em}
  \caption{{Synthetic NASCAR$^{\circledR}$,}
    an example of Bayesian inference in a recurrent
    switching linear dynamical system (rSLDS).
    {\bf(a)} In this case,
    the true dynamics switch between four states, causing the
    continuous latent state,~${x_t \in \reals^2}$, to trace ovals
    like a car on a NASCAR$^\circledR$ track. The dynamics of
    the most likely discrete state at a particular location are
    shown with arrows.
    {\bf (b)} The observations,~$y_t \in \reals^{10}$,
    are a linear projection with additive Gaussian noise (colors not given; for visualization only).
    {\bf (c)} The rSLDS correctly infers the continuous state trajectory,
    up to affine transformation. It also learns to partition the continuous
    space into discrete regions with different dynamics.
    {\bf (d)} Posterior samples of the discrete state sequence match
    the true discrete states, and show uncertainty near the change points.
    {\bf (e)} Generative samples from a standard SLDS differ dramatically
    from the true latent states in (a), since the run lengths in the
    SLDS are simple geometric random variables that are independent of the continuous
    state.
    {\bf (f)} In contrast, the rSLDS learns to generate states that shares the
    same periodic nature of the true model.
    }
  \label{fig:nascar}
  \vspace{-1em}
\end{figure*}
\section{Experiments}

We demonstrate the potential of recurrent dynamics in a variety
of settings. First, we consider a case in which the underlying
dynamics truly follow an rSLDS, which illustrates some of the
nuances involved in fitting these rich systems. With this
experience, we then apply these models to simulated data from a
canonical nonlinear dynamical system --~the Lorenz attractor~--
and find that its dynamics are well-approximated by an rSLDS.
Moreover, by leveraging the \polyagamma augmentation,
these nonlinear dynamics can even be recovered from
discrete time series with large swaths of missing data, as we
show with a Bernoulli-Lorenz model. Finally, we apply these
recurrent models to real trajectories on basketball players
and discover interpretable, location-dependent behavioral states.

\subsection{Synthetic NASCAR$^\circledR$}

We begin with a toy example in which the true dynamics trace out
ovals, like a stock car on a NASCAR$^{\circledR}$
track.\footnote{Unlike real NASCAR drivers, these states turn right.}
There are four discrete states,~${z_t \in \{1,\ldots,4\}}$, that govern the
dynamics of a two dimensional continuous latent state,~${x_t \in
  \reals^2}$. Fig.~\ref{fig:nascar}a shows the dynamics of the most
likely state for each point in latent space, along with a sampled
trajectory from this system. The observations,~${y_t \in \reals^{10}}$
are a linear projection of the latent state with additive Gaussian
noise. The 10 dimensions of~$y_t$ are superimposed in Fig.~\ref{fig:nascar}b.
We simulated~$T=10^4$ time-steps of data and fit an rSLDS to these data
with~$10^3$ iterations of Gibbs sampling.

Fig.~\ref{fig:nascar}c shows a sample of the inferred latent state
and its dynamics. It recovers the four states and a rotated oval
track, which is expected since the latent states are non-identifiable
up to invertible transformation. Fig.~\ref{fig:nascar}d plots the
samples of~$z_{1:1000}$ as a function of Gibbs iteration, illustrating
the uncertainty near the change-points. 

From a decoding perspective, both the SLDS and the rSLDS are capable
of discovering the discrete latent states; however, the rSLDS is a
much more effective generative model. Whereas the standard SLDS
has only a Markov model for the discrete states, and hence generates
the geometrically distributed state durations in Fig~\ref{fig:nascar}e,
the rSLDS leverages the location of the latent state to govern the discrete dynamics
and generates the much more realistic, periodic data in Fig.~\ref{fig:nascar}f.

\subsection{Lorenz Attractor}
\begin{figure*}[t]
  \centering \includegraphics[width=6.5in]{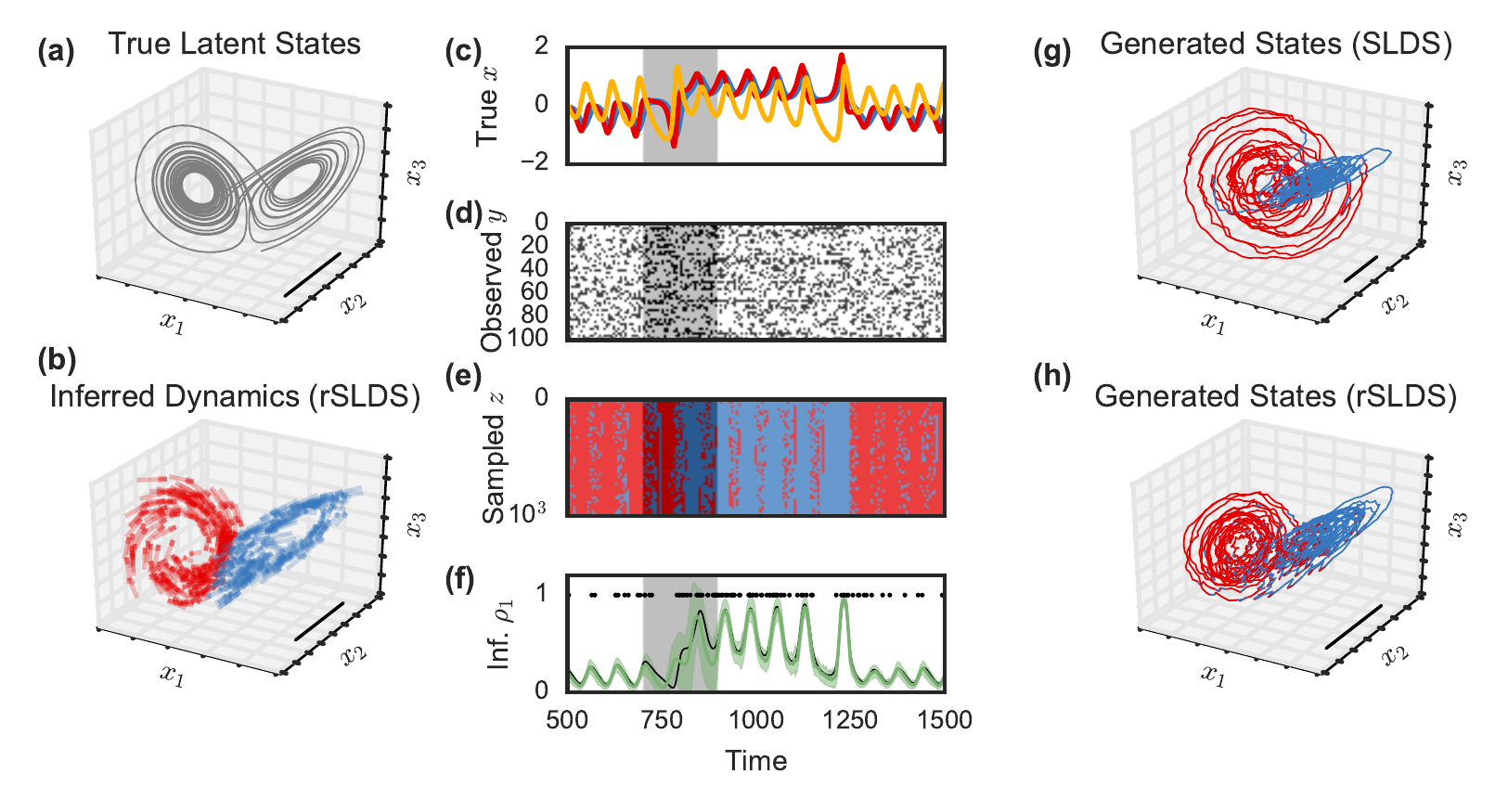}
  \vspace{-1em}
  \caption{A recurrent
    switching linear dynamical system (rSLDS) applied to simulated
    data from a Lorenz attractor --- a canonical nonlinear dynamical
    system. 
    {\bf(a)} The Lorenz attractor chaotically oscillates between two
    planes. Scale bar shared between (a), (b), (g) and (h).
    {\bf (b)} The rSLDS, with $x_t \in \reals^3$, identifies these two
    modes and their approximately linear dynamics, up to an invertible
    transformation.  It divides the space
    in half with a linear hyperplane.
    {\bf (c)} Unrolled over time, we see the points at which the Lorenz system
    switches from one plane to the other. Gray window denotes masked region
    of the data.
    \textbf{(d)} The observations come from a generalized linear
    model with Bernoulli observations and a logistic link function.
    \textbf{(e)} Samples of the discrete state show that the rSLDS
    correctly identifies the switching time even in the missing data.
    \textbf{(f)} The inferred probabilities (green) for the first output dimension
    along with the true event times (black dots) and the true
    probabilities (black line).  Error
    bars denote~${\pm 3}$ standard deviations under posterior.
    {\bf (g)} Generative samples from a standard SLDS differ substantially
    from the true states in {(a)} and are quite unstable.
    {\bf (h)}~In contrast, the rSLDS learns to generate state sequences that closely
    resemble those of the Lorenz attractor.
    }
  \label{fig:lorenz}
  \vspace{-1em}
\end{figure*}

Switching linear dynamical systems offer a tractable approximation
to complicated nonlinear dynamical systems. Indeed, one of the principal
motivations for these models is that once they have been fit, we can
leverage decades of research on optimal filtering, smoothing, and control
for linear systems. However, as we show in the case of the Lorenz attractor,
the standard SLDS is often a poor generative model, and hence has difficulty
interpolating over missing data. The recurrent SLDS
remedies this by connecting discrete and continuous states.

Fig.~\ref{fig:lorenz}a shows the states of a Lorenz attractor,
whose nonlinear dynamics are given by,
\begin{align}
  \frac{\mathrm{d} \bx}{\mathrm{d} t} &=
  \begin{bmatrix}
    \alpha (x_2 - x_1) \\
    x_1 ( \beta - x_3) - x_2 \\
    x_1 x_2 - \gamma x_3
  \end{bmatrix}.
\end{align}
Though nonlinear and chaotic, we see that the Lorenz attractor
roughly traces out ellipses in two opposing planes. Fig.~\ref{fig:lorenz}c
unrolls these dynamics over time, where the periodic nature and the
discrete jumps become clear.

Rather than directly observing the Lorenz states,~$x_{1:T}$, we simulate~${N=100}$
dimensional discrete observations from a generalized linear
model,
\begin{align}
  \rho_{t,n} &= \sigma(c_n^\trans x_t + d_n), & 
  y_{t,n} &\sim \distBernoulli(\rho_{t,n}),
\end{align}
where~${\sigma(\cdot)}$ is the logistic
function. A window of observations is shown in Fig.~\ref{fig:lorenz}d.
Just as we leveraged the \polyagamma augmentation to
render the continuous latent states conjugate with the multinomial discrete
state samples, we again leverage the augmentation scheme to
render them conjugate with Bernoulli observations.
As a further challenge, we also hold out a slice of data for~${t \in [700,900)}$,
identified by a gray mask in the center panels.
We provide more details in the supplementary material.


\begin{figure*}[t]
  \centering
  \includegraphics[width=6.6in]{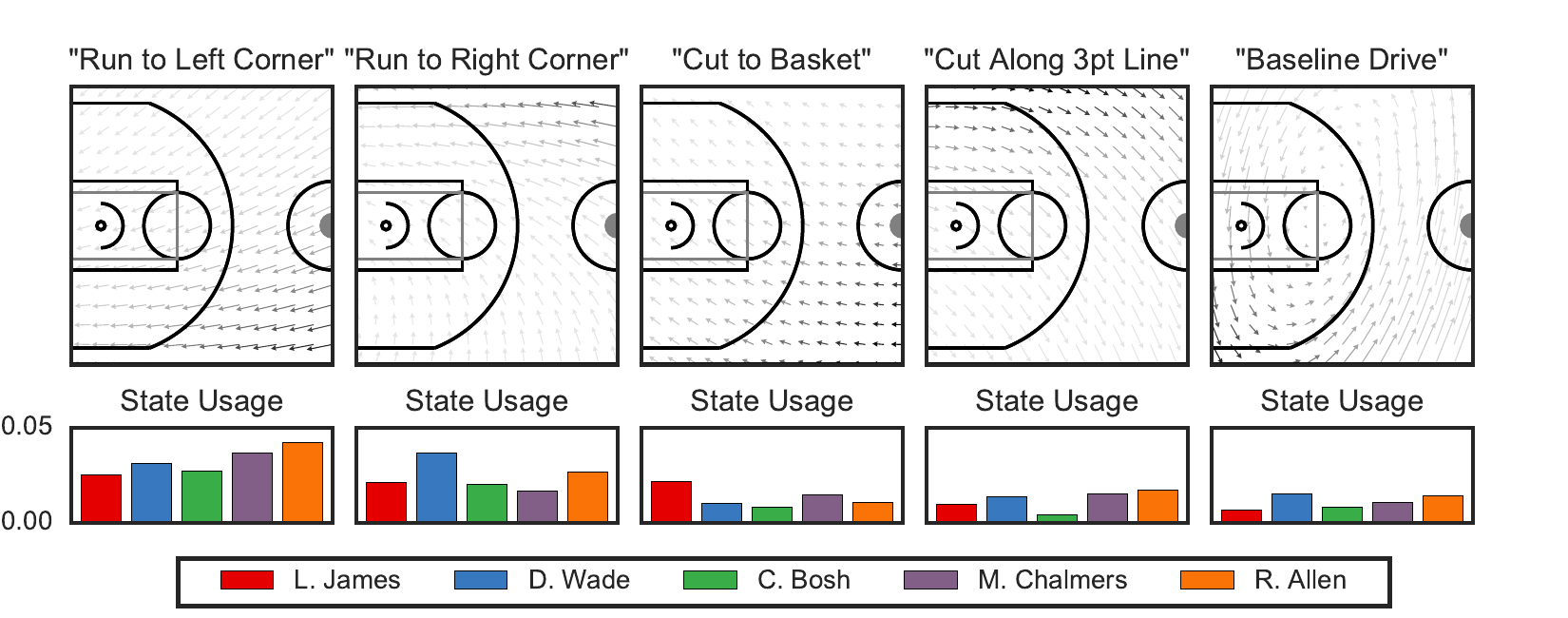}
  \vspace{-1em}
  \caption{Exploratory analysis of NBA player trajectories from the
    Nov. 1, 2013 game between the Miami Heat and the Brooklyn Nets.
    \textbf{(Top)} When applied to trajectories
    of five Heat players, the recurrent AR-HMM (ro) discovers~${K=30}$ discrete states with linear
    dynamics; five hand-picked states are shown here along with our names.
    Speed of motion is proportional to length of arrow. Location-dependent 
    state probability is proportional to opacity of the arrow.
    \textbf{(Bottom)} The probability with which each player uses
    the corresponding state under the posterior.
    }
  \label{fig:bball}
  \vspace{-1em}
\end{figure*}

Fitting an rSLDS via the same procedure described above, we
find that the model separates these two planes into two distinct
states, each with linear, rotational dynamics shown in Fig.~\ref{fig:lorenz}b.
Note that the latent states are only identifiable up to invertible
transformation.
Comparing Fig.~\ref{fig:lorenz}e to panel c, we see that the rSLDS
samples changes in discrete state at the points of large jumps in the data,
but when the observations are masked, there is more uncertainty.
This uncertainty in discrete state is propagated to uncertainty in the
event probability,~$\rho$, which is shown for the first output dimension
in Fig.~\ref{fig:lorenz}f. The times~${\{t: y_{t,1}=1\}}$ are shown as dots,
and the mean posterior probability~$\bbE[\rho_{t,1}]$ is shown with~$\pm 3$ standard
deviations.  Again, even in
the absence of observations, the model is capable of intelligently
interpolating, inferring when there should be a discrete change in dynamics.

Finally, as expected, data generated from a standard SLDS fit in the
same manner is quite different from the real data. It switches
from one state to another independent of the~$x$ location, resulting in large divergences
from the origin. In contrast, the rSLDS states are noisy yet qualitatively similar to those of the true Lorenz attractor.

\subsection{Basketball Player Trajectories}

We further illustrate our recurrent models with an application to
the trajectories run by five National Basketball Association (NBA)
players from the Miami Heat in a game against the Brooklyn Nets on
Nov.~1st, 2013. We are given trajectories,~${y^{(p)}_{1:T_p} \in \reals^{T_p \times 2}}$,
for each player~$p$.
We treat these trajectories as independent realizations of a ``recurrence-only''
AR-HMM with a shared set of~${K=30}$ states.
Positions are recorded every 40ms; combining the five players
yields 256,103 time steps in total. We use our rAR-HMM to
discover discrete dynamical states as well as the court locations in
which those states are most likely to be deployed.
We fit the model with 200 iteration of Gibbs sampling,
initialized with a draw from the prior. 

The dynamics of five of the discovered
states are shown in Fig.~\ref{fig:bball} (top), along with the names
we have assigned them. Below, we show the frequency with which each
player uses the states under the posterior distribution. 
First, we notice lateral symmetry; some
players drive to the left corner whereas others drive to the right.
Anecdotally, Ray Allen is known to shoot more from the left corner,
which agrees with the state usage here. 
The third state corresponds to a drive toward the hoop, which 
is most frequently used by LeBron James. Other states correspond
to unique plays made by the players, like cuts along the three-point line
and drives along the baseline. The complete set of states is
shown in the supplementary material.

\section{Discussion}
Through a variety of experiments, we have demonstrated how the
addition of recurrent connections from continuous states to
future discrete states lends increased flexibility and interpretability to the
classical switching linear dynamical system. This work is
similar in spirit to the \emph{piecewise affine} (PWA) framework
in control systems \citep{sontag1981nonlinear}. While Bayesian inference algorithms
have been derived for the special case of the rAR-HMM \citep{juloski2005bayesian},
the class of recurrent state space models has eluded a general
Bayesian treatment. However,
heuristic methods for learning
PWA models mimic our initialization procedure \citep{paoletti2007identification}.



The recurrent SLDS models strike a balance between flexibility
and tractability. Composing linear systems and linear partitions
achieves globally nonlinear dynamics while admitting efficient
Bayesian inference algorithms.  Directly modeling nonlinearities in the
dynamics and transition models, e.g. via Gaussian
processes or recurrent neural networks (RNNs), either necessitates more complex
non-conjugate inference and learning algorithms~\citep{frigola2013bayesian}
or non-Bayesian, data-intensive gradient methods~\citep{hochreiter1997long,
  graves2013generating} (though see \citet{gal2015theoretically} for
recent steps toward Bayesian learning for RNNs). Moreover, specifying
a reasonable class of nonlinear dynamics models may be more challenging
than for conditionally linear systems.
Finally, while these augmentation schemes do not apply to nonlinear
models, recent developments in structured variational autoencoders
\citep{johnson2016structured} provide new tools to extend the rSLDS
with nonlinear transition or emission models.

Recurrent state space models provide a flexible and interpretable
class of models for many nonlinear time series. Our Bernoulli-Lorenz example is
promising for other discrete domains, like multi-neuronal
spike trains with globally nonlinear dynamics~\citep[e.g.][]{sussillo2016lfads}.
Likewise, beyond the realm of basketball,
these models naturally apply to modeling of consumer shopping behavior ---
e.g. how do environmental features influence shopper paths? --- and
could be extended to model social behavior in multiagent systems.
These are exciting avenues for future work.

\subsubsection*{Acknowledgments}
SWL is supported by the Simons Foundation SCGB-418011.
ACM is supported by the Applied Mathematics Program within the Office of Science Advanced Scientific Computing Research of the U.S. Department of Energy under contract No. DE-AC02-05CH11231.
RPA is supported by NSF IIS-1421780 and the Alfred P. Sloan Foundation.
DMB is supported by NSF IIS-1247664, ONR N00014-11-1-0651, DARPA FA8750-14-2-0009, DARPA N66001-15-C-4032, Adobe, and the Sloan Foundation.
LP is supported by the Simons Foundation SCGB-325171; DARPA N66001-15-C-4032; ONR N00014-16-1-2176; IARPA MICRONS D16PC00003.

\bibliographystyle{plainnat}
\bibliography{refs}

\clearpage
\appendix

\section{Stochastic Variational Inference}
The main paper introduces a Gibbs sampling algorithm for the
recurrent SLDS and its siblings, but it is straightforward to
derive a mean field variational inference algorithm as well.
From this, 
we can immediately derive a stochastic variational inference (SVI)
\citep{hoffman2013stochastic} algorithm for conditionally
independent time series.

We use a structured mean field approximation on the augmented model,
\begin{multline*}
  p(z_{1:T}, x_{1:T}, \omega_{1:T}, \theta \given y_{1:T}) \\
  \approx q(z_{1:T}) \, q(x_{1:T}) \, q(\omega_{1:T}) \, q(\theta; \eta).
\end{multline*}
The first three factors will not be explicitly parameterized;
rather, as with Gibbs sampling, we leverage standard message passing
algorithms to compute the necessary expectations with respect to these
factors. Moreover,~$q(\omega_{1:T})$ further factorizes as,
\begin{align*}
  q(\omega_{1:T}) &= \prod_{t=1}^T \prod_{k=1}^{K-1} q(\omega_{t,k}).
\end{align*}
To be concrete, we also expand the parameter factor,
\begin{multline*}
  q(\theta; \eta) = \prod_{k=1}^K
  q(R_k, r_k \given \eta_{\mathsf{rec},k}) 
  \, q(A_k, b_k, B_k; \eta_{\mathsf{dyn},k}) \\
  \times q(C_k, d_k, D_k; \eta_{\mathsf{obs},k}).
\end{multline*}

The algorithm proceeds by alternating between optimizing $q(x_{1:T})$,
$q(z_{1:T})$,~$q(\omega_{1:T})$, and~$q(\theta)$.

\paragraph{Updating~$q(x_{1:T})$.}

Fixing the factor on the discrete states $q(z_{1:T})$, the optimal variational
factor on the continuous states $q(x_{1:T})$ is determined by,
\begin{multline*}
  \ln q(x_{1:T}) = \ln \psi(x_1) + \sum_{t=1}^{T-1} \ln \psi(x_t, x_{t+1})\\
    + \sum_{t=1}^{T-1} \ln \psi(x_t, z_{t+1}, \omega_t)
    + \sum_{t=1}^T \ln \psi(x_t; y_t)
    + c.
  \notag
\end{multline*}
where
\begin{align}
  \psi(x_1) &= \E_{q(\theta)q(z)} \ln p(x_{1} \given z_1, \theta)
  \label{eq:lds_init_potential}
  \\
  \psi(x_t, x_{t+1}) &= \E_{q(\theta)q(z)} \ln p(x_{t+1} \given x_t, z_t, \theta),
  \label{eq:lds_pair_potential}
  \\
  \psi(x_t, z_{t+1}) &= \E_{q(\theta)q(z)q(\omega)} \ln p(z_{t+1} \given x_t, z_t, \omega_t, \theta),
  \label{eq:lds_rec_potential}
\end{align}
Because the densities $p(x_1 \given z_1, \theta)$ and $p(x_{t+1} \given x_t,
z_t, \theta)$ are Gaussian exponential families, the expectations in
Eqs.~\eqref{eq:lds_init_potential}-\eqref{eq:lds_pair_potential} can be
computed efficiently, yielding Gaussian potentials with natural parameters that
depend on both $q(\theta)$ and $q(z_{1:T})$.
Furthermore, each $\psi(x_t; y_t)$ is itself a Gaussian potential.
As in the Gibbs sampler, the only non-Gaussian potential comes from the logistic
stick breaking model, but once again, the \polyagamma augmentation scheme comes
to the rescue. After augmentation, the potential as a function of~$x_t$ is,
\begin{align*}
  \E_{q(\theta)q(z)q(\omega)} &\ln p(z_{t+1} \given x_t, z_t, \omega_t, \theta) \\
  &= -\frac{1}{2} \nu_{t+1}^\trans \Omega_t \nu_{t+1} + \nu_{t+1}^\trans \kappa(z_{t+1})  + c.
\end{align*}
Since~${\nu_{t+1} = R_{z_t} x_t + r_{z_t}}$ is linear in~$x_t$, this is another
Gaussian potential. As with the dynamics and observation potentials, the
recurrence weights,~${(R_k, r_k)}$, also have matrix normal
factors, which are conjugate after augmentation.
We also need access
to~$\bbE_q[\omega_{t,k}]$; we discuss this computation below.

After augmentation, the
overall factor $q(x_{1:T})$ is a Gaussian linear dynamical system with natural
parameters computed from the variational factor on the dynamical parameters
$q(\theta)$, the variational parameter on the discrete states $q(z_{1:T})$,
the recurrence potentials~$\{\psi(x_t, z_t, z_{t+1})\}_{t=1}^{T-1}$, and
the observation model potentials $\{\psi(x_t; y_t)\}_{t=1}^T$.

Because the optimal factor $q(x_{1:T})$ is a Gaussian linear dynamical system,
we can use message passing to perform efficient inference.
In particular, the expected sufficient statistics of $q(x_{1:T})$ needed for
updating $q(z_{1:T})$ can be computed efficiently.

\paragraph{Updating~$q(\omega_{1:T})$.}
We have,
\begin{align*}
  &\ln q(\omega_{t,k}) = \bbE_q \ln p(z_{t+1} \given \omega_t, x_t) + c \\
  &= -\frac{1}{2} \bbE_q[ \nu_{t+1}^2]  \omega_{t,k} \\
  &\quad + \bbE_{q(z_{1:T})} \ln p_{\distPolyaGamma}(\omega_{t,k} \given \bbI[z_{t+1} \geq k], 0) + c
\end{align*}
While the expectation with respect to~$q(z_{1:T})$ makes this
challenging, we can approximate it with a sample,~$\hat{z}_{1:T} \sim q(z_{1:T})$.
Given a fixed value~$\hat{z}_{1:T}$ we have,
\begin{align*}
  q(\omega_{t,k}) &= p_{\distPolyaGamma}(\omega_{t,k} \given \bbI[\hat{z}_{t+1} \geq k], \bbE_q[ \nu_{t+1}^2]).
\end{align*}
The expected value of the distribution is available in closed form:
\begin{align*}
\bbE_q[\omega_{t,k}] &= \frac{\bbI[\hat{z}_{t+1} \geq k]}{2 \bbE_q[ \nu_{t+1}^2]}
\tanh \left(\tfrac{1}{2} \bbE_q[ \nu_{t+1}^2] \right).
\end{align*}

\paragraph{Updating~$q(z_{1:T})$.}
Similarly, fixing $q(x_{1:T})$ the optimal factor $q(z_{1:T})$
is proportional to
\begin{equation}
  \exp \left\{
    \ln \psi(z_1)
    + \sum_{t=1}^{T-1} \ln \psi(z_t, x_t, z_{t+1})
    + \sum_{t=1}^T \ln \psi(z_t)
  \right\},
  \notag
\end{equation}
where
\begin{align*}
  &\psi(z_1) = \E_{q(\theta)} \ln p(z_1 \given \theta) + \E_{q(\theta)q(x)} \ln p(x_1 \given z_1, \theta)\\
  &\psi(z_t,x_t, z_{t+1}) = \E_{q(\theta) q(x_{1:T})} \ln p(z_{t+1} \given z_t, x_t)
  \\
  &\psi(z_t)= \E_{q(\theta)q(x)} \ln p(x_{t+1} \given x_t, z_t, \theta)
\end{align*}
The first and third densities are exponential families; these
expectations can be computed efficiently. The challenge is the recurrence
potential,
\begin{align*}
  \psi(z_t, x_t, z_{t+1}) &= \bbE_{q(\theta), q(x)} \ln \pisb(\nu_{t+1}).
\end{align*}
Since this is not available in closed form, we approximate this expectation
with Monte Carlo over~$x_t$,~$R_k$, and~$r_k$. 
The resulting factor $q(z_{1:T})$ is an HMM with natural parameters that are
functions of $q(\theta)$ and $q(x_{1:T})$, and
the expected sufficient statistics required for updating $q(x_{1:T})$
can be computed efficiently by message passing in the same manner.

\paragraph{Updating~$q(\theta)$.}
To compute the expected sufficient statistics for the mean field update on
$\eta$, we can also use message passing, this time in both
factors $q(x_{1:T})$ and $q(z_{1:T})$ separately.
The required expected sufficient statistics are of the form
\begin{gather}
  \E_{q(z)} \I[z_t=i, z_{t+1}=j], \quad \E_{q(z)} \I[z_t = i],
  \notag
  \\
  \E_{q(z)} \I[z_t = k] \E_{q(x)} \! \left[ x_t x_{t+1}^\T \right],
  \\
  \E_{q(z)} \I[z_t = k] \E_{q(x)} \! \left[ x_t x_t^\T \right],
  \quad
  \E_{q(z)}  \I[z_1 = k] \E_{q(x)} \! \left[ x_1 \right],
  \notag
\end{gather}
where $\I[\, \cdot \, ]$ denotes an indicator function.
Each of these can be computed easily from the marginals $q(x_t, x_{t+1})$ and
$q(z_t, z_{t+1})$ for $t=1,2,\ldots,T-1$, and these marginals can be computed
in terms of the respective graphical model messages.

Given the conjugacy of the augmented model, the dynamics and observation factors will
be MNIW distributions as well. These allow closed form expressions for
the required expectations,
\begin{align*}
  \bbE_q[A_k], & & \bbE_q[b_k], & & \bbE_q[A_k B_k^{-1}], & & \bbE_q[b_k B_k^{-1}], & & \bbE_q[B_k^{-1}],  \\
  \bbE_q[C_k], & & \bbE_q[d_k], & & \bbE_q[C_k D_k^{-1}], & & \bbE_q[d_k D_k^{-1}], & & \bbE_q[D_k^{-1}].  
\end{align*}
Likewise, the conjugate matrix normal prior on~${(R_k, r_k)}$ provides access to
\begin{align*}
  \bbE_q[R_k], & & \bbE_q[R_k R_k^\trans], & & \bbE_q[r_k].
\end{align*}

\paragraph{Stochastic Variational Inference.}
Given multiple, conditionally independent observations of time
series,~${\{y_{1:T_p}^{(p)}\}_{p=1}^P}$ (using the same notation
as in the basketball experiment), it is straightforward to derive
a stochastic variational inference (SVI) algorithm \citep{hoffman2013stochastic}.
In each iteration, we sample a random time series;
run message passing to compute the optimal local factors,~$q(z_{1:T_p}^{(p)})$,
~$q(x_{1:T_p}^{(p)})$, and~$q(\omega_{1:T_p}^{(p)})$; and then
use expectations with respect to these local factors as unbiased
estimates of expectations with respect to the complete dataset
when updating the global parameter factor,~$q(\theta)$. Given
a single, long time series, we can still derive efficient SVI
algorithms that use subsets of the data,
as long as we are willing to accept minor, controllable bias \citep{johnson2014stochastic, foti2014stochastic}.

\section{Stick Breaking and Decision Lists}
As mentioned in Section~4, one of the less desirable
features of the logistic stick breaking regression model is its
dependence on the ordering of the output dimensions; in our case, on the
permutation of the discrete states~$\{1, 2, \ldots, K\}$. To alleviate
this issue, we first do a greedy search over permutations by fitting
a decision list to~${(x_t, z_t), z_{t+1}}$ pairs. A decision list is
an iterative classifier of the form,
\begin{align*}
  z_{t+1} = \begin{cases}
    o_1 & \text{if } \bbI[p_1] \\
    o_2 & \text{if } \bbI[\neg p_1 \wedge p_2] \\
    o_3 & \text{if } \bbI[\neg p_1 \wedge \neg p_2 \wedge p_3] \\
    \vdots \\
    o_K & \text{o.w.},
    \end{cases}
\end{align*}
where~$\left(o_1, \ldots, o_K \right)$ is a permutation of~$(1, \ldots, K)$,
and~$p_1, \ldots, p_k$ are predicates that depend on~$(x_t, z_t)$ and evaluate
to true or false. In our case, these predicates are given by logistic functions,
\begin{align*}
  p_j &= \sigma(r_j^\trans x_t) > 0. 
\end{align*}

We fit the decision list using a greedy approach: to determine~$o_1$
and~$r_1$, we use maximum a posterior estimation to fit logistic
regressions for each of the~$K$ possible output values. For the
~$k$-th logistic regression, the inputs are~$x_{1:T}$ and the outputs
are~${y_t=\bbI[z_{t+1}=k]}$.  We choose the best logistic regression
(measured by log likelihood) as the first output. Then we remove those
time points for which~$z_{t+1}=o_1$ from the dataset and repeat,
fitting~$K-1$ logistic regressions in order to determine the second
output,~$o_2$, and so on.

After iterating through all~$K$ outputs, we have a permutation
of the discrete states. Moreover, the predicates~$\{r_k\}_{k=1}^{K-1}$
serve as an initialization for the recurrence weights,~$R$, in our
model.

\section{Bernoulli-Lorenz Details}
The \polyagamma augmentation makes it easy to handle discrete observations
in the rSLDS, as illustrated in the Bernoulli-Lorenz experiment. 
Since the Bernoulli likelihood is given by,
\begin{align*}
  p(y_t \given z_t, \theta) &= \prod_{n=1}^N \distBernoulli(\sigma(c_n^\trans x_t + d_n) \\
  &= \prod_{n=1}^N \frac{(e^{c_n^\trans x_t + d_n})^{y_{t,n}}}{1+e^{c_n^\trans x_t + d_n}},
\end{align*}
we see that it matches the form of~(7) with,
\begin{align*}
  \nu_{t,n} &= c_n^\trans x_t + d_n, & 
  b(y_{t,n}) &= 1, & 
  \kappa(y_{t,n}) &= y_{t,n} - \frac{1}{2}. 
\end{align*}
Thus, we introduce an additional set of \polyagamma auxiliary variables,
\begin{align*}
  \xi_{t,n} &\sim \distPolyaGamma(1, 0),
\end{align*}
to render the model conjugate. Given these auxiliary variables, the
observation potential is proportional to a Gaussian distribution on~$x_t$,
\begin{align*}
  \psi(x_t, y_t) &\propto \distNormal(Cx_t + d \given \Xi_t^{-1} \kappa(y_t), \Xi_t^{-1}),
\end{align*}
with
\begin{align*}
  \Xi_t &= \diag \left([\xi_{t,1}, \ldots, \xi_{t,N}] \right), \\
  \kappa(y_t) &= [\kappa(y_{t,1}), \ldots, \kappa(y_{t,N})].
\end{align*}
Again, this admits efficient message passing inference for~$x_{1:T}$.
In order to update the auxiliary variables, we sample from their conditional
distribution,~${\xi_{t,n} \sim \distPolyaGamma(1, \nu_{t,n})}$.

This augmentation scheme also works for binomial, negative binomial, and
multinomial obsevations as well \citep{polson2013bayesian}.

\section{Basketball Details}
\begin{figure*}[t]
  \centering
  \includegraphics[width=6.6in]{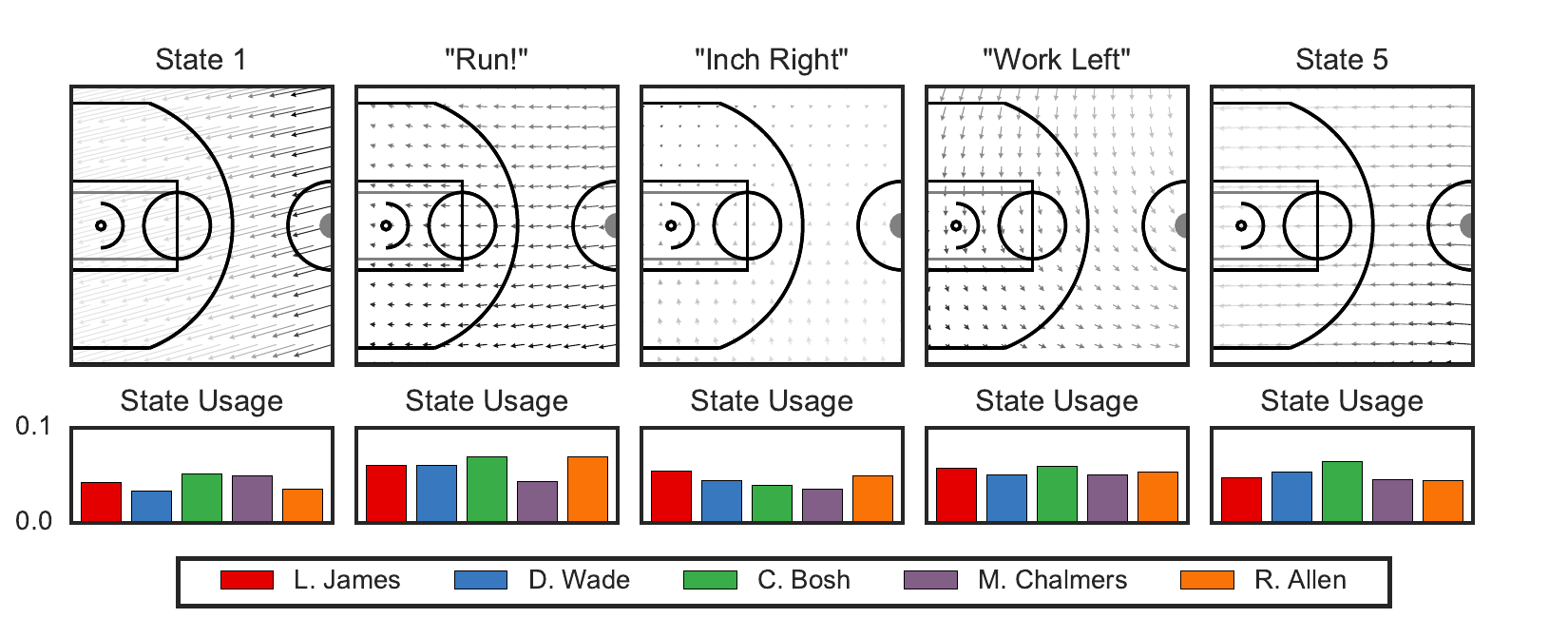}\\
  \includegraphics[width=6.6in]{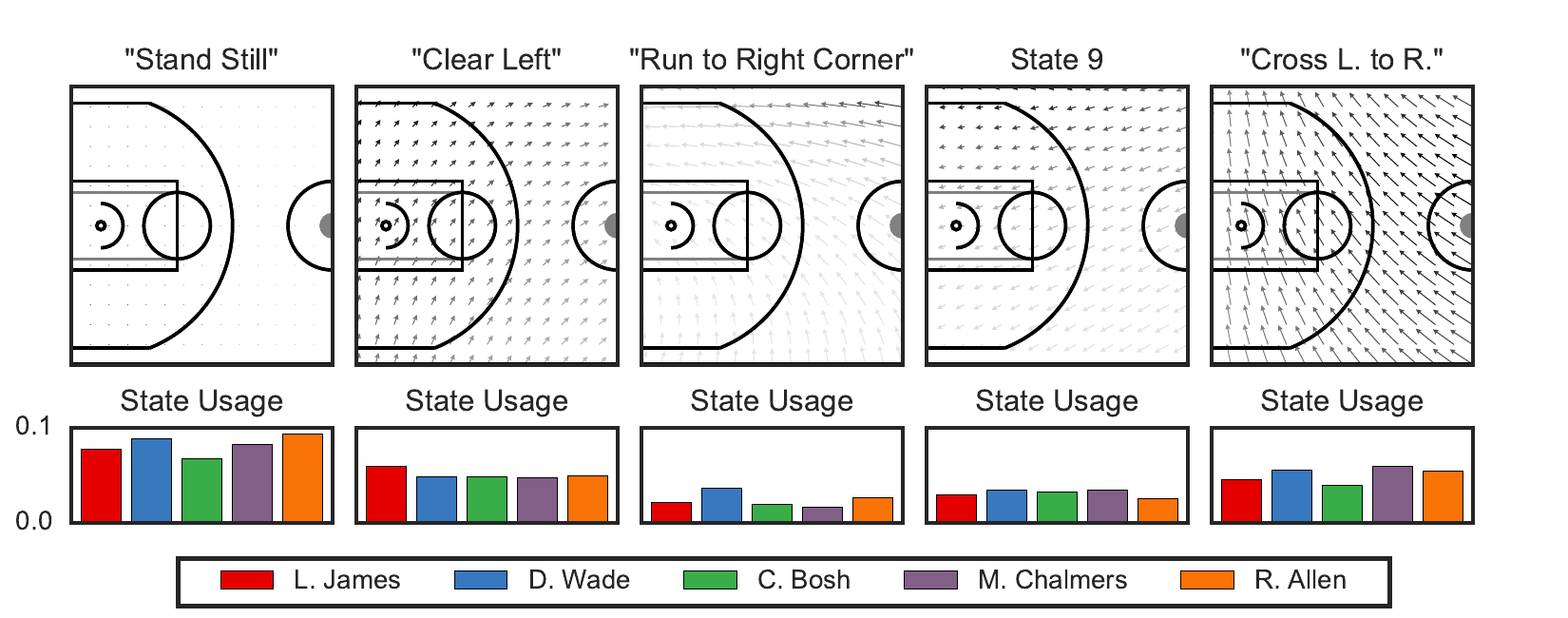}\\
  \includegraphics[width=6.6in]{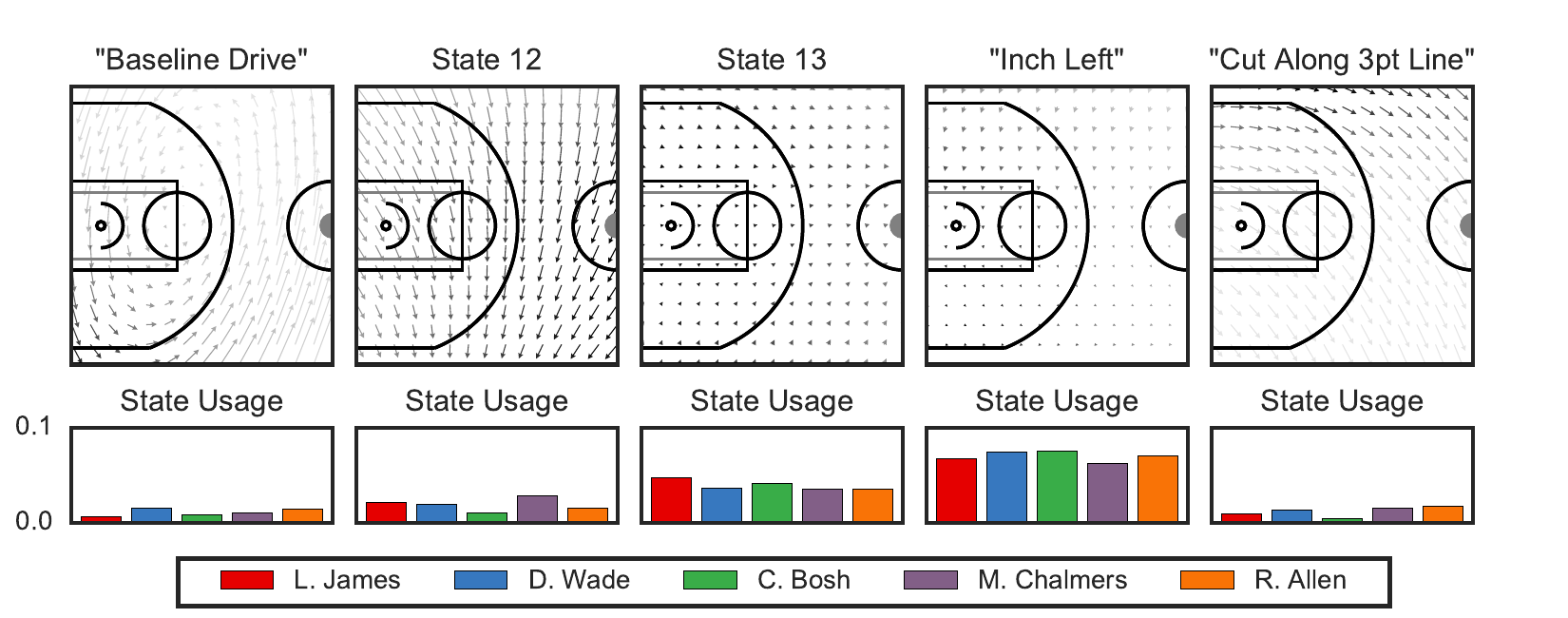}\\
  \caption{All of the inferred basketball states
    }
  \label{fig:bball_all1}
\end{figure*}

\begin{figure*}[t]
  \centering
  \includegraphics[width=6.6in]{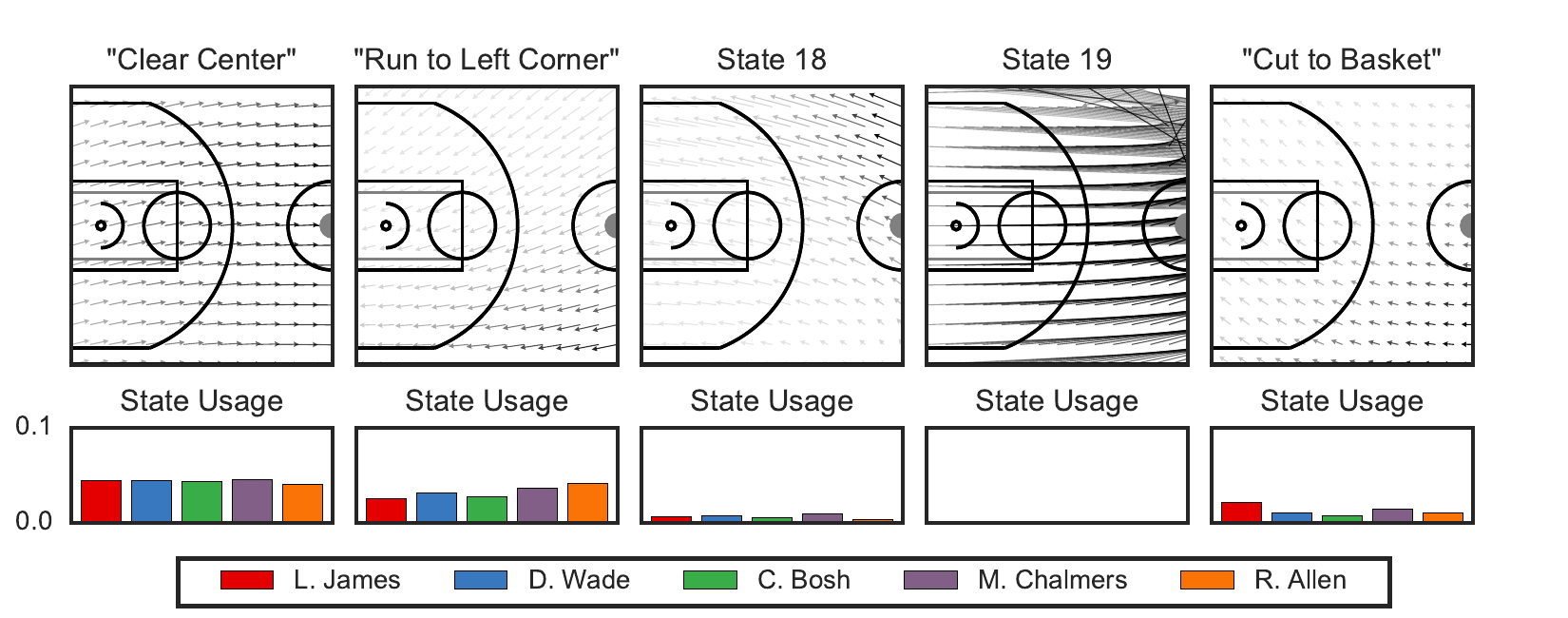}\\
  \includegraphics[width=6.6in]{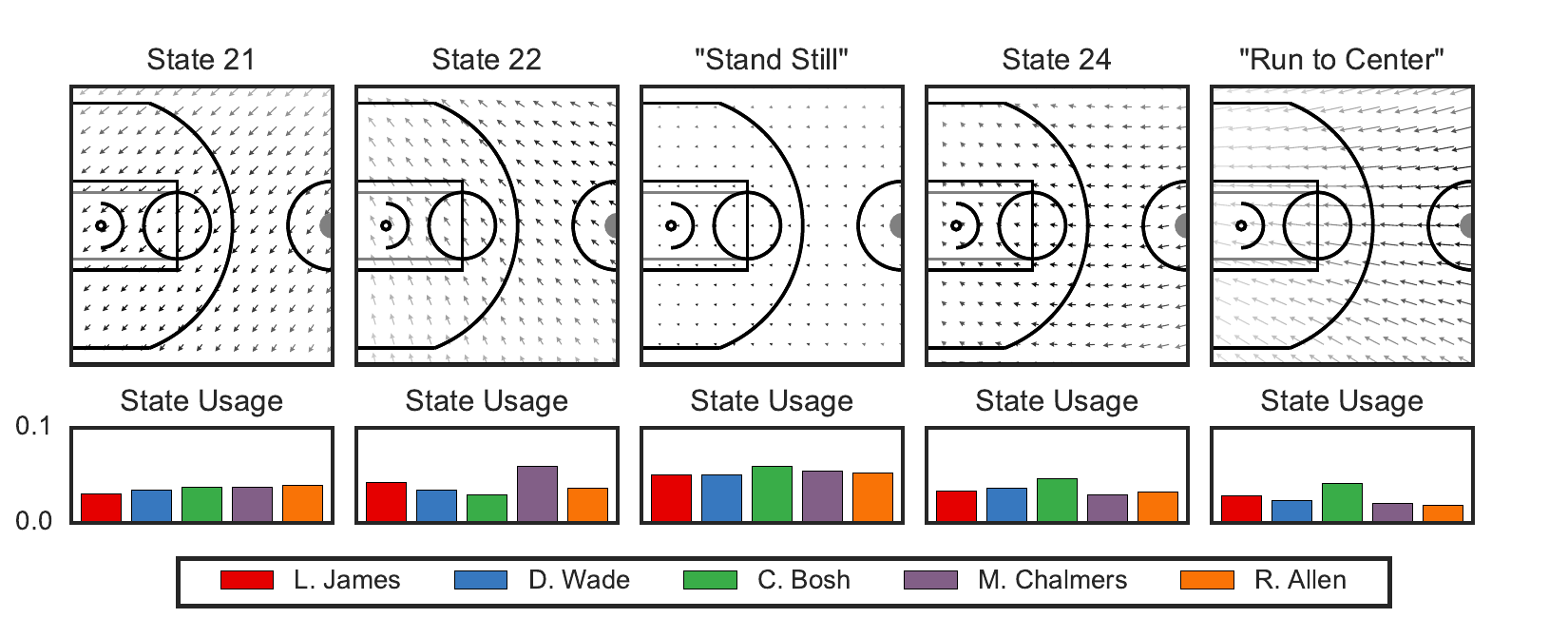}\\
  \includegraphics[width=6.6in]{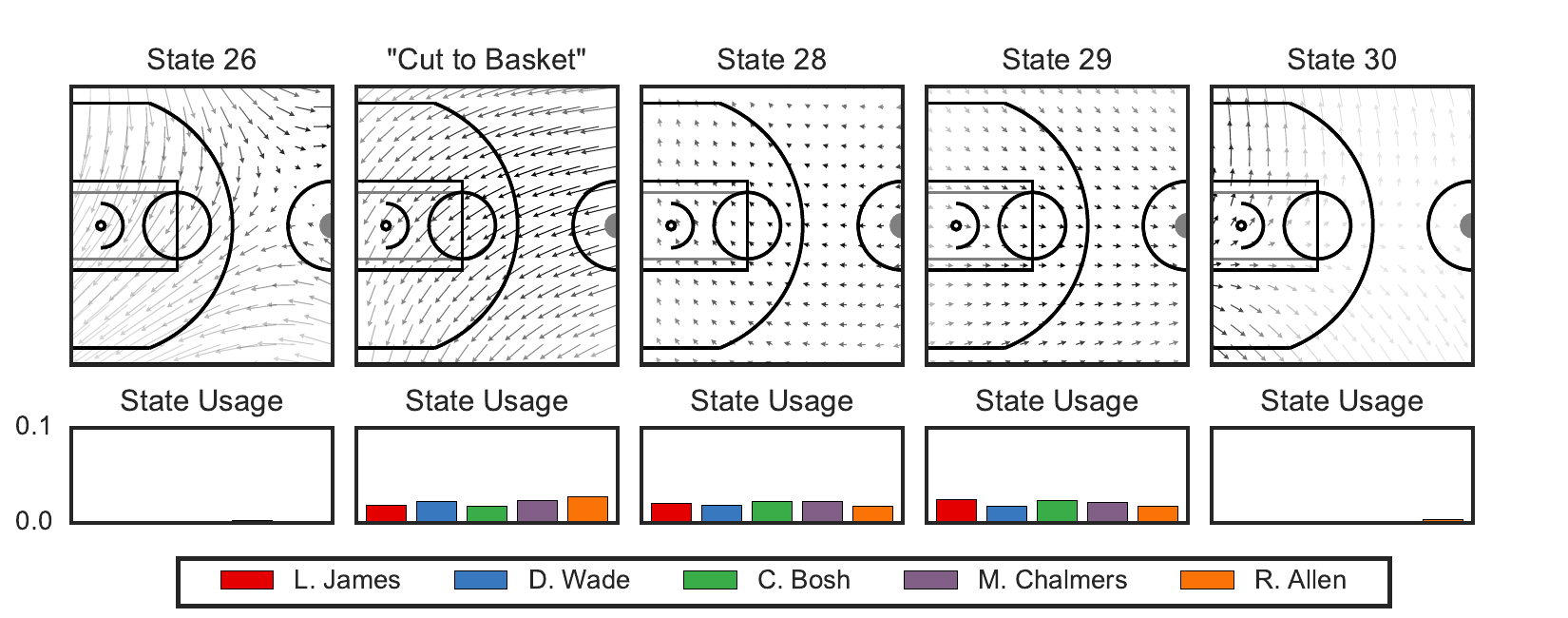}
  \caption{All of the inferred basketball states
    }
  \label{fig:bball_all2}
\end{figure*}

For completeness, Figures~\ref{fig:bball_all1} and~\ref{fig:bball_all2} show all~${K=30}$ inferred
states of the rAR-HMM (ro) for the basketball data.

\end{document}